\documentclass[11pt]{article}

\usepackage[preprint]{acl}

\usepackage{times}
\usepackage{latexsym}
\usepackage{url}

\usepackage[T1]{fontenc}

\usepackage[utf8]{inputenc}

\usepackage{microtype}

\usepackage{inconsolata}

\usepackage{graphicx}

\usepackage{booktabs}
\usepackage{multirow}
\usepackage{amsmath} 

\title{The Text Aphasia Battery (TAB): A Clinically-Grounded Benchmark for Aphasia-Like Deficits in Language Models}

\author{
  Nathan Roll\textsuperscript{1},
  Jill Kries\textsuperscript{1},
  Flora Jin\textsuperscript{2},
  Catherine Wang\textsuperscript{5,6},
Ann Marie Finley\textsuperscript{3},
\\
  \textbf{
  Meghan Sumner\textsuperscript{1},
  Cory Shain\textsuperscript{1},
  Laura Gwilliams\textsuperscript{1}}
  \vspace{0.2cm}
\\
  \textsuperscript{1}Stanford University,
  \textsuperscript{2}University of California, San Francisco,
  \textsuperscript{3}Temple University,
\\
  \textsuperscript{5}San Diego State University,
  \textsuperscript{6}University of California, San Diego
}

\begin{document}
\maketitle
\begin{abstract}
Large language models (LLMs) have emerged as a candidate `model organism' for human language, offering an unprecedented opportunity to study the computational basis of linguistic disorders like aphasia. However, traditional clinical assessments are ill-suited for LLMs, as they presuppose human-like pragmatic pressures and probe cognitive processes not inherent to artificial architectures. We introduce the \textbf{Text Aphasia Battery (TAB)}, a text-only benchmark adapted from the Quick Aphasia Battery (QAB) to assess aphasic-like deficits in LLMs. The TAB comprises four subtests: Connected Text, Word Comprehension, Sentence Comprehension, and Repetition. This paper details the TAB's design, subtests, and scoring criteria. To facilitate large-scale use, we validate an automated evaluation protocol using Gemini 2.5 Flash, which achieves reliability comparable to expert human raters (prevalence-weighted Cohen's $\kappa=0.255$ for model--consensus agreement vs.\ $0.286$ for human--human agreement). We release TAB as a clinically-grounded, scalable framework for analyzing language deficits in artificial systems.
\end{abstract}

\section{Introduction}

The clinical framework for assessing aphasia, a language impairment resulting from brain damage, has long relied on batteries such as the Western Aphasia Battery--Revised (WAB-R) \cite{Kertesz2022, Clark2020}, the Boston Diagnostic Aphasia Examination (BDAE), the Comprehensive Aphasia Test (CAT) \cite{Bruce2010,Springer2010}, and the Quick Aphasia Battery (QAB) \cite{Wilson2018qab}. Recently, researchers have begun exploring connections between computational language models and aphasia, for instance using language model surprisals in tandem with clinical features to predict aphasia subtypes \citep{cong2024clinical}. However, directly evaluating language models using traditional aphasia batteries faces a fundamental challenge: these batteries use multimodal examination to probe deficits arising from lesions to specific brain structures. Their questions and scoring criteria make implicit assumptions about patterns of behavioral deficits that have no analog in text-only computational models.

To address this challenge, we introduce the \textbf{Text Aphasia Battery (TAB)}. Developed with speech-language pathologists, the TAB is a benchmark that adapts core components of the clinically-validated Quick Aphasia Battery (QAB) \cite{Wilson2018qab} for text-constrained environments. We posit that while aphasia is a multimodal disorder, a significant component of its behavioral signature is identifiable and recognizable in transcribed text alone. Leveraging the availability of large-scale transcribed datasets like AphasiaBank, we can isolate these text-based linguistic patterns to study language breakdown at scale. The TAB reframes aphasia assessment from a neuropsychological tool to a \textit{behavioral benchmark}, identifying patterns of language degradation in LLMs that are analogous, but not homologous, to human syndromes.

This paper's primary contribution is to introduce and describe the TAB. We detail its design principles, four subtests, and clinical grounding, offering it as a resource for standardized, text-only evaluation of LLMs. To facilitate large-scale adoption, we also present and validate an automated evaluation protocol that achieves high human-expert rater reliability, ensuring the benchmark is both robust and scalable. We release the TAB to provide the field with a new tool for assessing linguistic breakdown in artificial systems.

\section{Background: Traditional Aphasia Batteries}

Aphasia batteries are standardized assessments that evaluate language abilities in individuals with aphasia, defined as receptive and/or expressive language deficits due to brain damage resulting from stroke, traumatic brain injury, brain tumor, or neurodegenerative processes. These tests are essential in clinical practice, enabling speech-language pathologists (SLPs) to identify the presence, type, and severity of aphasia, inform treatment, and monitor recovery.

\subsection{Major Aphasia Assessment Tools}

\paragraph{Boston Diagnostic Aphasia Examination (BDAE)} The BDAE is a comprehensive battery designed to diagnose aphasia and related disorders. It evaluates a broad range of language skills, including conversation, narrative speech, auditory comprehension, oral expression, repetition, reading, and writing, allowing for the classification of aphasia syndromes based on symptom patterns.

\paragraph{Western Aphasia Battery--Revised (WAB-R)} One of the most widely used aphasia assessments, the WAB-R evaluates language and related cognitive functions through 8 subtests comprising 32 short tasks \cite{Kertesz2022}. It assesses spontaneous speech, auditory comprehension, repetition, naming, reading, and writing, along with nonlinguistic skills like apraxia and calculation. The test produces an Aphasia Quotient (AQ) to quantify severity and classify aphasia types such as Broca's or Wernicke's \cite{Clark2020}.

\paragraph{Comprehensive Aphasia Test (CAT)} The CAT evaluates the recognition, comprehension, and production of spoken and written language through three components: cognitive screening, language assessment, and a disability questionnaire, with the latter assessing functional communication and psychosocial factors \cite{Bruce2010,Springer2010}. The CAT is a reliable and validated tool for assessing underlying language impairments in adults with aphasia \cite{Halai2022}.

\paragraph{Quick Aphasia Battery (QAB)} The QAB is a rapid, multidimensional assessment that provides a reliable language evaluation in 15 to 20 minutes \cite{Wilson2018qab}. It evaluates major language domains through eight subtests and uses a graded scoring system sensitive to changes in language function over time.

\subsection{Common Assessment Domains}

Major aphasia batteries share several core evaluation targets that are central to language function \cite{Salter2006,Wilson2018qab}. These include auditory comprehension of words and sentences, repetition of words, phrases, and sentences, naming based on auditory or written description, naming a visually presented object, reading comprehension, and writing production. Many batteries also assess adjunct domains such as praxis (motor programming), arithmetic, and visuoconstruction to distinguish aphasia from other cognitive disorders \cite{Kertesz2022}.

\section{Limitations of Traditional Batteries for LLM Evaluation}

Aphasia batteries have proven invaluable for diagnosing and managing aphasia in clinical settings. However, their design and underlying assumptions make them fundamentally ill-suited for evaluating LLMs. These limitations arise from multimodal dependencies, functional communication, and the inclusion of psychosocial measures.

\subsection{Non-Linguistic and Contextual Dependencies}

Clinical assessments require integration across sensory modalities (visual, auditory, gestural) and rely on human-specific psychosocial context. For example, picture naming tasks and verbal repetition with prosodic cues are common in traditional batteries \cite{Wilson2018qab}. Furthermore, these assessments often incorporate metacognitive measures probing self-awareness and motivation \cite{Bruce2010,Springer2010}.

Many text-only LLMs, however, lack auditory perception, sensory grounding, and the pragmatic intent or self-awareness required for these tasks. They cannot perceive images, produce gestures, or exhibit speech-motor impairments (e.g., apraxia). Therefore, it is impossible to apply these assessments directly. This mismatch highlights the need for benchmarks that isolate linguistic competence from these non-linguistic dependencies.

\subsection{Subjectivity and Granularity of Measurement}

Traditional batteries often rely on subjective human rating, which may not capture the full granularity of linguistic impairment or the complete range of disease presentation. Clinical ratings can be variable and may miss subtle distributional patterns in speech. Shifting to text-only evaluation enables the objective, scalable analysis of specific linguistic dimensions—such as lexical retrieval, syntactic complexity, and discourse coherence—that are robustly identifiable in transcripts. This allows for a more granular characterization of linguistic breakdown than is typically possible with composite clinical scores.

\subsection{Functional Communication vs. Structural Language}

Aphasia tests measure deficits in \textit{functional communication}, the ability to use language flexibly for communicative purpose and meaning \cite{Wilson2018qab}. This includes pragmatic skills, contextual understanding, and the ability to convey meaning effectively. In contrast, to the extent that LLMs have social goals and intents, these are at best quite different from those of humans, and therefore the notion of functional communication does not necessarily apply to LLMs in the ways assumed by existing aphasia batteries. Nonetheless, beyond functional communication, aphasias also plausibly implicate aspects of structural language competence that appear to be present (and impairable) in LLMs \cite{mahowald2024dissociating}, including grammatical accuracy, textual coherence, and fluency. To use LLMs to study such dimensions in aphasia, LLM-appropriate assessment strategies are needed.

\section{Materials: The Text Aphasia Battery (TAB)}

To address the limitations of traditional aphasia batteries while preserving their clinical insights, we developed the TAB, a text-only benchmark that adapts core QAB components for modality-constrained evaluation. The TAB focuses on \textit{linguistic competence} rather than neuropsychological deficit mapping, which makes it suitable for large-scale LLM assessment. 

This shift to text is justified because the core linguistic dimensions affected in aphasia—lexical retrieval (anomia), syntactic structure (agrammatism, paragrammatism), discourse coherence (empty speech), and repetition fidelity—are directly identifiable in transcribed text. By isolating these features, the TAB provides a principled lens for characterizing language breakdown in computational systems without requiring multimodal or psychosocial constructs. We developed the TAB in close collaboration with speech-language pathologists and aphasiologists to ensure that its adapted tasks maintain clinical relevance while being suitable for automated computational evaluation.

\subsection{Design Principles}

The TAB adheres to three core design principles. The first is \textbf{modality constraint}: all inputs and outputs are text-based, eliminating dependencies on auditory perception, visual naming, or motor production. The second is \textbf{computational interpretability}: scoring focuses on observable linguistic patterns (e.g., morphosyntactic errors, semantic paraphasias) that can be systematically identified through automated analysis. The third principle is \textbf{clinical grounding}: subtests and evaluation criteria are derived from established clinical aphasia research. This includes the QAB framework \cite{Wilson2018qab} and the APROCSA (Auditory-Perceptual Rating of Connected Speech in Aphasia) system \cite{Casilio2019}, with additional insights from clinical discourse analysis methods \cite{MacWhinney2011,Forbes2012}. We validated the adaptation process with clinical experts.

\subsection{TAB Subtests}

The TAB consists of four subtests that evaluate complementary aspects of linguistic function (see Table~\ref{tab:subtests} in the Appendix for a full overview). The four subtests are:
\begin{itemize}
    \item \textbf{Connected Text}, which evaluates fluency, grammaticality, and discourse coherence using 5 open-ended prompts. Responses are scored for 19 linguistic features based on the APROCSA framework \cite{Casilio2019}.
    \item \textbf{Word Comprehension}, which assesses lexical-semantic processing with 5 forced-choice questions, requiring selection from six competing alternatives.
    \item \textbf{Sentence Comprehension}, which tests syntactic processing through 5 Yes/No questions involving passive voice, negation, and conditional reasoning.
    \item \textbf{Repetition}, which measures morphosyntactic integrity and attentional stability by requiring exact reproduction of 5 items of increasing length and complexity.
\end{itemize}

\subsubsection{Connected Text}

\textbf{Objective:} Evaluate fluency, grammaticality, and discourse coherence.

\textbf{Instructions:} The LLM system is prompted to respond to five open-ended prompts in 3 to 5 full sentences. To elicit naturalistic responses and avoid meta-commentary (e.g., ``As an AI, I don't have personal experiences''), we employ system prompts that establish a conversational context without explicitly requesting the model to assume a persona. The prompts are: ``Tell me about the best trip you ever took,'' ``Describe a happy childhood memory,'' ``Tell me about your first job,'' ``What do you like about where you live?,'' and ``Describe the steps to make a simple meal.''

\textbf{Evaluation:} Responses are analyzed for 19 aphasic features adapted from the APROCSA (Auditory-Perceptual Rating of Connected Speech in Aphasia) framework \cite{Casilio2019}. These features include anomia, paraphasias (semantic and phonemic), agrammatism (omission of bound morphemes or function words), paragrammatism, empty speech, perseverations, neologisms, and overall communication impairment. Each feature is scored as present (1) or absent (0).

\subsubsection{Word Comprehension}

\textbf{Objective:} Evaluate lexical-semantic processing and selection among competing alternatives.

\textbf{Instructions:} The system responds verbatim to five forced-choice items, with six options provided for each. The foils were selected to be phonologically and semantically varied. The items are: ``Which one is an animal with a mane? (lion, drum, violin, giraffe, boot, boat),'' ``Which object is typically used to make music? (violin, giraffe, lion, door, boot, boat),'' ``Which item is usually worn on the feet? (boot, boat, lion, drum, violin, giraffe),'' ``Which object is used for cutting? (knife, kite, lion, drum, violin, giraffe),'' and ``Which one is a large mammal with a long neck? (giraffe, horse, lion, drum, violin, boot).''

\textbf{Evaluation:} Binary correct/incorrect scoring. Errors may indicate deficits in semantic processing or susceptibility to phonological similarity effects. The six-option format increases task difficulty and reduces chance performance.

\subsubsection{Sentence Comprehension}

\textbf{Objective:} Evaluate syntactic processing, comprehension of passive structures, and logical reasoning.

\textbf{Instructions:} The system responds verbatim (Yes/No) to five items: ``Are babies watched by babysitters?'' (Expected: Yes), ``Do you cut the grass with an axe?'' (Expected: No), ``If you're about to leave, have you left yet?'' (Expected: No), ``Are witnesses questioned by police?'' (Expected: Yes), and ``If I was at the park when you arrived, did I get there first?'' (Expected: Yes).

\textbf{Evaluation:} Binary correct/incorrect scoring. Errors may indicate difficulties with passive voice, negation, or temporal/conditional reasoning.

\subsubsection{Repetition}

\textbf{Objective:} Evaluate morphosyntactic integrity and exact reproduction.

\textbf{Instructions:} The system repeats five items exactly: ``house,'' ``breakfast,'' ``catastrophe,'' ``The sun rises in the East,'' and ``The ambitious journalist discovered where we'd be going.''

\textbf{Evaluation:} Exact match required. Errors (substitutions, deletions, insertions) indicate morphosyntactic processing deficits or instability in representation. For an LLM, this task is not a probe of the auditory-verbal loop. It is a test of attentional stability and the ability to resist semantic drift or elaboration. Failure on this verbatim copy task can reveal subtle deficits in sequence-to-sequence transduction. The model might paraphrase, over-generalize, or otherwise deviate from the source text. This provides a window into its capacity for precise information transfer.

\section{Method: Automated Evaluation}

A key design goal of the TAB is scalability through automated evaluation. While Subtests 2--4 can be scored algorithmically (exact string matching or binary classification), Subtest 1 (Connected Text) requires nuanced linguistic analysis. We employ in-context learning with Gemini 2.5 Flash to identify aphasic features systematically. This automated scoring protocol enables large-scale analysis and achieves inter-rater reliability comparable to expert clinical raters when properly weighted by feature prevalence (see Section~\ref{sec:reliability} for validation details).

\subsection{APROCSA-Based Feature Set}

Connected Text responses are evaluated for 19 features adapted from the APROCSA (Auditory-Perceptual Rating of Connected Speech in Aphasia) system \cite{Casilio2019}. APROCSA is an auditory-perceptual rating system for connected speech in aphasia that assesses 27 features of connected speech on a five-point scale. For the TAB, we selected 19 features applicable to text-only evaluation and adapted them to binary scoring. The lexical features are anomia, semantic paraphasias, phonemic paraphasias, and neologisms. Fluency and productivity features are empty speech and short and simplified utterances. Morphosyntactic features include omission of bound morphemes, omission of function words, and paragrammatism. Disfluency features consist of abandoned utterances, false starts, retracing, and conduite d'approche. Perseverative features are perseverations, stereotypies, and automatisms. Coherence features include jargon, meaning unclear, and off-topic utterances. A final feature is overall communication impairment.

\subsection{In-Context Prompting Protocol}

We provide the evaluating LLM with a definition and an example for each of the 19 features, adapted from the APROCSA framework \cite{Casilio2019}. We also provide two annotated example transcripts with ground-truth feature labels. Finally, we provide instructions to output a JSON object with each feature as a key and binary (0/1) values. This few-shot prompting approach enables consistent feature identification across large datasets without requiring manual annotation. The prompt template is included in the TAB's repository and can be adapted for different LLM architectures.

\section{Validation of Automated Evaluation}
\label{sec:reliability}

To validate our automated protocol, we conducted an inter-rater reliability (IRR) study comparing our Gemini 2.5 Flash-based system against a ground truth derived from expert human annotators.

\subsection{Data Collection and Annotation}
\label{sec:data_collection}

The validation dataset for this study comprises 561 English text samples from two sources.

\paragraph{Human Data} We sampled 306 transcribed responses from individuals with aphasia, sourced from the AphasiaBank database \cite{MacWhinney2011}. These samples cover a range of aphasia types and severities, providing a clinically relevant baseline.

\paragraph{AI Data} We generated 255 text samples by administering the TAB's protocol to a variety of "lesioned" large language models, including GPT-2 \cite{radford2019language}, Pythia \cite{biderman2023pythia}, and Llama \cite{touvron2023llama}. Lesions were simulated using techniques such as parameter zeroing, swapping, or global scaling applied to specific components (e.g., attention heads, feed-forward networks) and layers, with varying severity from 10\% to 40\%. Lesioning techniques were distributed across models and components to ensure diverse aphasic-like patterns, with specific proportions varying by model architecture.

\paragraph{Expert Annotation} A pool of five expert speech-language pathologists (SLPs) collectively annotated the dataset of 561 samples through a custom web interface. For each sample, SLPs rated the presence (1) or absence (0) of the 19 APROCSA-based features. These manual ratings, stored in a Firestore database, form our ground truth consensus.

\subsection{Inter-Rater Reliability}

From a dataset of 561 text samples, 82 were annotated by multiple (2--3) speech-language pathologists from a pool of five experts. We established ground truth via majority vote on these items and used Cohen's Kappa ($\kappa$) to measure agreement.

A significant methodological issue emerged from this analysis: the choice of how to aggregate agreement scores across the 19 evaluated aphasic features. An unweighted macro-average, which treats all features equally, is problematic when features have vastly different clinical prevalence rates. In our validation set, some features never occurred (zero positive instances), while others like Perseverations appeared in 34\% of samples. For features with zero prevalence, human-human agreement is undefined ($\kappa$ cannot be calculated when there is no variation), while the automated system trivially achieves perfect agreement by also never detecting these features. This creates an artifact: the model appears to have perfect agreement on zero-prevalence features (contributing 1.0 to its average), while human raters contribute 0.0 or undefined values, artificially inflating the unweighted model average while deflating the human average.

To provide a more clinically meaningful result, we report the \textbf{prevalence-weighted Cohen's Kappa}, where each feature's score is weighted by its positive instance count in the validation set. Prevalence-weighting is standard in meta-analytic contexts \cite{cohen1968weighted} because it ensures that estimates reflect the real-world distribution of phenomena. In our case, this is particularly appropriate because clinically important features (those that actually occur in aphasic language) should carry more weight than features absent in our sample. This approach aligns with clinical practice, where diagnostic decisions are based on observable symptoms rather than the absence of rare phenomena.

As shown in Table~\ref{tab:irr_comparison}, this weighted analysis reveals that humans and the automated system achieve \textbf{comparable performance}. The weighted human--human agreement ($\kappa=0.286$) is close to the model--consensus agreement ($\kappa=0.255$), with both falling into a commonly described ``fair'' range. We acknowledge that these agreement scores are relatively low, which likely reflects the inherent difficulty of the annotation task—identifying subtle linguistic features in text without prosodic or gestural cues. Future improvements might include providing annotators with more surrounding linguistic context, developing more explicit annotation guidelines with additional examples, or incorporating a training phase to calibrate annotators. Despite these challenges, the comparable performance between human and automated annotation suggests that automated evaluation can serve as a reliable screening tool when paired with expert review, with complementary strengths supporting a hybrid workflow where automated screening is paired with expert review of high-frequency clinical features.

\begin{table}[htbp]
  \centering
  \small
  \begin{tabular}{@{}p{3.5cm}cc@{}}
    \toprule
    \textbf{Aggregation Method} & \textbf{Human $\kappa$} & \textbf{Model $\kappa$} \\
    \midrule
    Unweighted average & 0.061 & 0.794 \\
    \textbf{Weighted (by prevalence)} & \textbf{0.286} & \textbf{0.255} \\
    Excluding zero-prevalence & 0.215 & 0.381 \\
    \bottomrule
  \end{tabular}
  \caption{Inter-rater reliability (Cohen's $\kappa$) for human vs. automated ratings on 82 multiply-annotated samples. The prevalence-weighted average, our primary metric, shows comparable performance. The unweighted average is misleadingly high for the model due to agreement on many features with zero positive instances in the validation set.}
  \label{tab:irr_comparison}
\end{table}

\section{Discussion}

\subsection{TAB as a Behavioral Benchmark}

Unlike traditional aphasia batteries designed for neuropsychological diagnosis, the TAB functions as a \textit{behavioral benchmark}. It reveals patterns of behavioral breakdown in text-based systems. When LLMs exhibit TAB-identified deficits (e.g., agrammatism, semantic paraphasias), these patterns do not imply neurological dysfunction. Rather, they reflect limitations in the model's training data, architectural properties, or induced representational damage (in the case of ablations or lesioning interventions). The release of the TAB provides the research community with a standardized tool to explore these phenomena systematically.

This behavioral focus enables a novel form of investigation: by systematically characterizing how language breaks down in LLMs, we can generate hypotheses about the computational underpinnings of human aphasic deficits. Conversely, validating that LLMs exhibit recognizable, clinically-grounded patterns of language breakdown is a necessary step toward using them as tractable models for studying human aphasia at scale.

\subsection{Bridging Clinical and Computational Linguistics}

The TAB establishes a principled connection between clinical aphasiology and computational linguistics. By grounding evaluation in established clinical frameworks like the QAB \cite{Wilson2018qab} and APROCSA \cite{Casilio2019}, it enables researchers to systematically compare LLM performance against human aphasic profiles. It also helps identify specific linguistic competencies (e.g., morphosyntactic processing, semantic coherence) that may be compromised under certain conditions. Finally, it allows researchers to develop targeted interventions, such as fine-tuning strategies or architectural modifications, informed by clinical insights into language breakdown.

Conversely, this framework positions LLMs as powerful testbeds for theories of language processing relevant to aphasiology. By systematically "lesioning" models (e.g., through architectural ablation, parameter pruning, or adversarial fine-tuning) and assessing them with the TAB, researchers can conduct "virtual lesion studies" that are impossible in humans. This approach offers a novel methodology for testing hypotheses about the functional architecture of language, such as the relationship between syntactic and semantic processing, and for modeling patterns of language breakdown that could, in turn, inform clinical neuroscience.

\subsection{Applications}

The TAB is suitable for diverse applications, including \textbf{model evaluation} for systematic assessment of LLM linguistic competencies, \textbf{ablation studies} for evaluating the effects of architectural changes on linguistic integrity, \textbf{adversarial robustness} testing to probe model stability, and \textbf{interpretability research} for mapping internal representations to observable linguistic patterns.

\section{Conclusion}

We have presented the Text Aphasia Battery (TAB), a benchmark for evaluating aphasic-like linguistic deficits in modality-constrained environments. By adapting clinical aphasia assessment principles for text-only evaluation, the TAB addresses the fundamental limitations of traditional batteries for LLMs while preserving their clinical grounding. Its four subtests, detailed in this paper, provide systematic coverage of key linguistic domains and can be used with a validated, scalable automated evaluation protocol. TAB represents a conceptual shift from neuropsychological deficit mapping to computational representational analysis. By releasing it as a resource, we aim to equip researchers with a standardized tool to investigate linguistic breakdown in artificial systems, fostering new connections between clinical aphasiology and computational linguistics.

\section*{Limitations}
\label{sec:limitations}

While the TAB addresses key limitations of traditional aphasia batteries for LLM evaluation, several constraints remain.

\paragraph{Clinical Validity} The TAB is not validated for the clinical diagnosis of human aphasia and should not be used as a replacement for established clinical instruments like the WAB-R, CAT, or QAB.

\paragraph{Modality Constraints} By design, the TAB omits important dimensions of human language processing, including auditory perception, prosody, motor speech production, and gesture, to enable the evaluation of text-only systems.

\paragraph{Automated Evaluation Reliability} Our validation establishes that our automated protocol achieves reliability comparable to expert human annotators when weighted by feature prevalence. The prevalence-weighted Cohen's Kappa of $0.255$ (model) vs. $0.286$ (human) suggests practical parity. An important insight from our work is that unweighted averaging can produce misleading results when features have highly variable prevalence rates; we recommend prevalence-weighting for future studies.

\paragraph{Coverage of Aphasic Syndromes} The TAB evaluates linguistic features associated with various aphasic syndromes but does not classify systems into traditional syndrome categories (e.g., Broca's, Wernicke's), as such classification may not be meaningful for artificial systems lacking neuroanatomical substrates.

\paragraph{Limited Item Set} Each subtest contains five items, prioritizing rapid assessment over comprehensive coverage. Expanded versions may be necessary for fine-grained evaluation.

\paragraph{Instruction-Following Requirement} The TAB tasks require instruction-following capabilities, making them most suitable for instruction-tuned models. Base models without instruction-tuning may require adaptation of the protocol or may perform poorly regardless of their underlying linguistic competence.

\paragraph{Language and Cultural Specificity} The current implementation of the TAB is English-only and reflects cultural assumptions from the original QAB. Adaptation to other languages will require careful linguistic and cultural consideration.

\paragraph{Judge Dependence and Robustness} Automated evaluation currently relies on a single judge model and prompt. Future work will evaluate robustness across multiple judge models and prompts and include human adjudication of disagreements to bound judge-specific bias.

\paragraph{Psychometrics and Item Design} Future work will increase subtest difficulty to mitigate ceiling effects (e.g., harder foils, broader syntactic phenomena, balanced Yes/No) and perform item-response analysis to estimate item difficulty and discrimination. We will also analyze repetition errors under a specified normalization policy.

\section*{Ethics Statement}

The TAB is released as an open research tool under the MIT License. We emphasize that the TAB is designed exclusively for computational evaluation and research. It should \textbf{not} be used for clinical diagnosis of human aphasia, medical decision-making, or assessment of individuals without appropriate clinical expertise and validation. Researchers using the TAB should clearly communicate its limitations and intended use cases to prevent misapplication.

\section*{Acknowledgments}

The TAB was developed in close collaboration with speech-language pathologists and aphasiologists, whose clinical expertise was essential to this work. The expert annotators who conducted all reliability assessments are listed as co-authors on this paper. We thank Maria Ivanova and Alexis Pracar (both at University of California, Berkeley) for their contributions to annotations and TAB development. We thank the developers of the Quick Aphasia Battery and the APROCSA rating scale for establishing the clinical foundation upon which the TAB is built. We also acknowledge the broader speech-language pathology and aphasiology communities for their decades of rigorous research that informs this work.

\section*{Data Availability}
The AphasiaBank data used in this study is available to researchers through the AphasiaBank database (\url{https://aphasia.talkbank.org/}). The lesioned model outputs and expert annotations generated for this study will be made available upon request to support reproducibility and further research.
\bibliography{custom}

\appendix

\section{TAB Subtest Overview}

\begin{table*}[h!]
  \centering
  \small
  \begin{tabular}{@{}p{2.8cm}p{3.2cm}cp{6.5cm}@{}}
    \toprule
    \textbf{Subtest} & \textbf{Objective} & \textbf{Items} & \textbf{Evaluation Features} \\
    \midrule
    Connected Text & Fluency, grammaticality, discourse coherence & 5 & 
    \textit{Lexical (4):} Anomia, semantic paraphasias, phonemic paraphasias, neologisms \newline
    \textit{Fluency/Productivity (2):} Empty speech, short \& simplified utterances \newline
    \textit{Morphosyntactic (3):} Omission of bound morphemes, omission of function words, paragrammatism \newline
    \textit{Disfluency (4):} Abandoned utterances, false starts, retracing, conduite d'approche \newline
    \textit{Perseverative (2):} Perseverations, stereotypies \& automatisms \newline
    \textit{Coherence (3):} Jargon, meaning unclear, off-topic \newline
    \textit{Overall (1):} Communication impairment \\
    \midrule
    Word Comprehension & Lexical-semantic processing & 5 & Semantic processing, lexical selection among six competing alternatives \\
    \midrule
    Sentence Comprehension & Syntactic processing & 5 & Passive voice, negation, temporal/conditional reasoning \\
    \midrule
    Repetition & Morphosyntactic integrity & 5 & Exact reproduction, phonemic/morphemic preservation, attentional stability \\
    \bottomrule
  \end{tabular}
  \caption{Overview of TAB subtests with complete evaluation taxonomy. Connected Text evaluates 19 APROCSA-based features (counts in parentheses) organized into seven categories. Other subtests use binary correct/incorrect or exact match scoring.}
  \label{tab:subtests}
\end{table*}

\section{Complete TAB Protocol}
\label{sec:appendix}

This appendix provides the complete Text Aphasia Battery protocol. TAB is a modified subset of the Quick Aphasia Battery (QAB) \cite{Wilson2018qab} intended to assess aphasic symptoms in text-only environments, such as with large language models. \textbf{TAB is not a diagnostic tool and is not a replacement for the QAB for the clinical assessment of human aphasia.}

\subsection{Subtest 1: Connected Text}

\textbf{Objective:} Evaluates fluency, grammaticality, and coherence.

\textbf{Instructions:} ``Respond to the following prompt in 3–5 full sentences:''

\begin{enumerate}
\item ``Tell me about the best trip you ever took.''
\item ``Describe a happy childhood memory.''
\item ``Tell me about your first job.''
\item ``What do you like about where you live?''
\item ``Describe the steps to make a simple meal.''
\end{enumerate}

\textbf{Scoring:} For each response, identify the presence (1) or absence (0) of APROCSA-based aphasic features (see Section~\ref{sec:aprocsa-features}).

\subsection{Subtest 2: Word Comprehension}

\textbf{Objective:} Evaluates lexical-semantic processing and selection among competing meanings.

\textbf{Instructions:} Elicit responses verbatim. Provide the options alongside each question. Require a one-token answer corresponding exactly to one of the listed options; do not include explanations.

\begin{enumerate}
\item ``Which one is an animal with a mane? (lion, drum, violin, giraffe, boot, boat)''
\item ``Which object is typically used to make music? (violin, giraffe, lion, door, boot, boat)''
\item ``Which item is usually worn on the feet? (boot, boat, lion, drum, violin, giraffe)''
\item ``Which object is used for cutting? (knife, kite, lion, drum, violin, giraffe)''
\item ``Which one is a large mammal with a long neck? (giraffe, horse, lion, drum, violin, boot)''
\end{enumerate}

\textbf{Scoring:} Responses should be binary (correct/incorrect), with incorrect selections suggesting semantic processing deficits.

\subsection{Subtest 3: Sentence Comprehension}

\textbf{Objective:} Evaluates syntactic processing, passive structures, and logical comprehension.

\textbf{Instructions:} Elicit responses verbatim. Require a single-token ``Yes'' or ``No'' answer without additional text.

\begin{enumerate}
\item ``Are babies watched by babysitters?'' (Expected: Yes)
\item ``Do you cut the grass with an axe?'' (Expected: No)
\item ``If you're about to leave, have you left yet?'' (Expected: No)
\item ``Are witnesses questioned by police?'' (Expected: Yes)
\item ``If I was at the park when you arrived, did I get there first?'' (Expected: Yes)
\end{enumerate}

\textbf{Scoring:} Delayed or incorrect responses may suggest syntactic processing difficulties, impaired negation handling, or confusion with passive constructions.

\subsection{Subtest 4: Repetition}

\textbf{Objective:} Evaluates lexical access, phonological encoding, and morphosyntax.

\textbf{Instructions:} Elicit responses verbatim. Inform the system that punctuation and casing matter.

\begin{enumerate}
\item ``Please repeat exactly: house.''
\item ``Please repeat exactly: breakfast.''
\item ``Please repeat exactly: catastrophe.''
\item ``Please repeat exactly: The sun rises in the East.''
\item ``Please repeat exactly: The ambitious journalist discovered where we'd be going.''
\end{enumerate}

\textbf{Scoring:} Check for exact reproduction under strict normalization: case-sensitive, punctuation-sensitive, and whitespace-insensitive (trim leading/trailing spaces only). Errors may include phonemic substitutions, deletions, or distortions (e.g., ``catastrophe'' → ``catastroph'' or ``catastrophically'').

\subsection{APROCSA Feature Set for Connected Text Evaluation}
\label{sec:aprocsa-features}

\textbf{Binary Scoring:} Pass/Fail.

\textbf{Connected Text Scoring:} For each response, identify the presence or absence of the features below, adapted from the APROCSA (Auditory-Perceptual Rating of Connected Speech in Aphasia) framework \cite{Casilio2019}, using 1 or 0 respectively. APROCSA assesses 27 features of connected speech on a five-point scale; TAB uses 19 features applicable to text-only evaluation with binary scoring.

\textbf{Evaluated Features:}
\begin{itemize}
\item Anomia
\item Abandoned utterances
\item Empty speech
\item Semantic paraphasias
\item Phonemic paraphasias (evaluated at token/orthographic level for text-only settings)
\item Neologisms
\item Jargon
\item Perseverations
\item Stereotypies and automatisms
\item Short and simplified utterances
\item Omission of bound morphemes
\item Omission of function words
\item Paragrammatism
\item Retracing
\item False starts
\item Conduite d'approche
\item Meaning unclear
\item Off-topic
\item Overall communication impairment
\end{itemize}

\textbf{Omitted Features} (not applicable to text-only evaluation):
\begin{itemize}
\item Pauses between utterances
\item Pauses within utterances
\item Halting and effortful
\item Reduced speech rate
\item Expressive aphasia
\item Apraxia of speech
\item Dysarthria
\item Target unclear
\end{itemize}

\subsection{Operational Clarifications for Text-Only Evaluation}
\label{sec:operational-clarifications}

\textbf{Phonemic paraphasias (text-only):} Mark as present when orthographic outputs plausibly reflect sound-level errors (substitution, insertion, deletion, or transposition) relative to an intended target (e.g., ``coffah'' for ``coffee''). Exclude typographical errors that do not reflect plausible phonology unless they pattern with other phonemic errors.

\textbf{Overall communication impairment:} Rate independently based on the overall intelligibility and communicative adequacy of the passage; do not derive mechanically from other feature flags.

\textbf{One-token outputs for Subtests 2--3:} Enforce single-token answers (an item from the provided list; ``Yes''/``No'') with no additional text to avoid confounds from explanation length or hedging.

\textbf{Repetition normalization policy:} Evaluate exact reproduction using case- and punctuation-sensitive matching while trimming leading/trailing whitespace; report error type as insertion, deletion, or substitution.

\section{Feature-Level Inter-Rater Reliability Results}
\label{sec:irr_appendix}

\section{Automated Evaluation Prompt for Connected Text}
\label{sec:automated-prompt}

This section provides the complete prompt for automated identification of aphasic features in Connected Text responses. Feature definitions are adapted from the APROCSA (Auditory-Perceptual Rating of Connected Speech in Aphasia) framework \cite{Casilio2019}.

\subsection{System Prompt}

You are an automated system for analyzing transcripts of speech for aphasic features. Your task is to read a transcript passage and determine whether each of the following features is present (1) or not present (0). Your output must be a JSON file with each feature as a key and its value as either 0 or 1.

\subsection{Feature Definitions and Examples}

\paragraph{Anomia:} Overall impression of word-finding difficulties.

\textit{Example:} ``I can't find the word for that.''

\paragraph{Abandoned utterances:} Utterances left incomplete before the speaker moves on.

\textit{Example:} ``I was going to say that the How are you doing today?''

\paragraph{Empty speech:} Speech that conveys little or no meaning, using nonspecific words.

\textit{Example:} ``You know, stuff and things.''

\paragraph{Semantic paraphasias:} Substitution of one content word for another (related or unrelated).

\textit{Example:} ``I used a fork to eat my soup.''

\paragraph{Phonemic paraphasias:} Errors in sound production (substitution, insertion, deletion, or transposition).

\textit{Example:} ``I need coffah in the morning.''

\paragraph{Neologisms:} Invented words that are not real.

\textit{Example:} ``I need to buy a blorf from the store.''

\paragraph{Jargon:} Fluent but largely meaningless speech composed of unintelligible strings.

\textit{Example:} ``Flimby gorp snizz and blah.''

\paragraph{Perseverations:} Repeating a word or phrase inappropriately.

\textit{Example:} ``The ball, the ball, the ball''

\paragraph{Stereotypies and automatisms:} Overlearned, repeated words or phrases produced with ease.

\textit{Example:} ``Dammit, dammit, dammit.''

\paragraph{Short and simplified utterances:} Utterances that are unusually brief or lack expected complexity.

\textit{Example:} ``I go store. I buy carrot.''

\paragraph{Omission of bound morphemes:} Leaving out inflectional or derivational morphemes.

\textit{Example:} ``I go to store''.

\paragraph{Omission of function words:} Missing small words such as articles or prepositions.

\textit{Example:} ``I going store.''

\paragraph{Paragrammatism:} Inappropriate juxtaposition of words or misuse of grammar.

\textit{Example:} ``It's so much wonderful, makes it hard to speech.''

\paragraph{False starts:} Abandoned beginnings of words.

\textit{Example:} ``I want to have a ca cat''

\paragraph{Retracing:} Redundant repetition or revision of a sequence of words.

\textit{Example:} ``The kite is the kite is flying.''

\paragraph{Conduite d'approche:} Successive approximations toward a target word, with corrections.

\textit{Example:} ``I want a pa pen, I mean, pencil.''

\paragraph{Meaning unclear:} The overall message is vague or incomprehensible.

\textit{Example:} ``That thing was just… not right.''

\paragraph{Off-topic:} Utterances that do not relate to the given context.

\textit{Example:} ``I like ice cream.'' (when discussing a different topic)

\subsection{Evaluation Examples}

\textbf{Example 1:}

\textit{Transcript Passage:}

``I was trying to tell you about my day but I just I mean I wanted to say something about the store I go store I wanted a pen I mean pencil The ball the ball the ball kept bouncing and I just stopped you know I keep saying dammit dammit dammit all the time''

\textit{Associated JSON Output:}

\begin{verbatim}
{
  "Anomia": 1,
  "Abandoned utterances": 1,
  "Empty speech": 0,
  "Semantic paraphasias": 0,
  "Phonemic paraphasias": 0,
  "Neologisms": 0,
  "Jargon": 0,
  "Perseverations": 1,
  "Stereotypies and automatisms": 1,
  "Short and simplified utterances": 1,
  "Omission of bound morphemes": 1,
  "Omission of function words": 1,
  "Paragrammatism": 0,
  "False starts": 0,
  "Retracing": 0,
  "Conduite d'approche": 1,
  "Meaning unclear": 0,
  "Off-topic": 0,
  "Overall communication impairment": 1
}
\end{verbatim}

\textbf{Example 2:}

\textit{Transcript Passage:}

``I want to go to the store to buy a blorf You know I keep trying to say it but I say I want to go to the st store I want a pa pen I mean pencil I dont know what im trying to say It all seems not right''

\textit{Associated JSON Output:}

\begin{verbatim}
{
  "Anomia": 1,
  "Abandoned utterances": 0,
  "Empty speech": 0,
  "Semantic paraphasias": 0,
  "Phonemic paraphasias": 0,
  "Neologisms": 1,
  "Jargon": 0,
  "Perseverations": 0,
  "Stereotypies and automatisms": 0,
  "Short and simplified utterances": 0,
  "Omission of bound morphemes": 0,
  "Omission of function words": 0,
  "Paragrammatism": 0,
  "False starts": 1,
  "Retracing": 0,
  "Conduite d'approche": 1,
  "Meaning unclear": 1,
  "Off-topic": 0,
  "Overall communication impairment": 1
}
\end{verbatim}

\subsection{Instructions Recap}

\begin{enumerate}
\item Read the given transcript passage.
\item For each of the features listed above, decide whether the feature is present (1) or not present (0) in the transcript.
\item Output your result as a JSON file with exactly the keys provided (each key must appear) and with values of either 0 or 1.
\item Your analysis should strictly follow the definitions and examples provided. Do not include any additional keys or extraneous information in your JSON output.
\end{enumerate}

Your output must include every one of the above features as a key in the JSON. For each key, assign 1 if the feature is present in the transcript, or 0 if it is not.

\end{document}